\documentclass[12pt,a4paper]{article}

\usepackage{arxiv}

\usepackage[utf8]{inputenc} 
\usepackage[T1]{fontenc}    
\usepackage{hyperref}       
\usepackage{url}            
\usepackage{booktabs}       
\usepackage{amsfonts}       
\usepackage{nicefrac}       
\usepackage{microtype}      
\usepackage{lipsum}
\usepackage{graphicx}
\usepackage{float}
\usepackage{amsmath}

\title{Fourier Transform Approach to \\Machine Learning II: Fourier Clustering  \\}

\author{
  Soheil Mehrabkhani \\
\\
  \texttt{soheil.mehrabkhani@alumni.uni-heidelberg.de} \\}

\begin{document}
\maketitle

\begin{abstract}

We propose a Fourier-based approach for optimization of several clustering algorithms. Mathematically, clusters data can be described by a density function represented by the Dirac mixture distribution. The density function can be smoothed by applying the Fourier transform and a Gaussian filter. The determination of the optimal standard deviation of the Gaussian filter will be accomplished by the use of a convergence criterion related to the correlation between the smoothed and the original density functions. In principle, the optimal smoothed density function exhibits local maxima, which correspond to the cluster centroids. Thus, the complex task of finding the centroids of the clusters is simplified by the detection of the peaks of the smoothed density function. A multiple sliding windows procedure is used to detect the peaks. The remarkable accuracy of the proposed algorithm demonstrates its capability as a reliable general method for enhancement of the clustering performance, its global optimization and also removing the initialization problem in many clustering methods. 

\end{abstract}


\vskip 0.5in

\section{Introduction}

The performance of several successful wide-spread clustering algorithms like K-Means [1-3] and Gaussian mixture method [1-3] depends on the assumption of the number of clusters and the position of their centroids. The a priori information of the number of clusters is a fundamental assumption for many clustering methods and without such information, the clustering computational complexity increases. Furthermore, the well-known initialization problem affects the performance and the global optimization of some clustering methods [2]. \\
The challenging task of the acquisition of a priori information can be converted to the much simpler task of peak detection if we use the concept of the density function described by the Dirac mixture distribution applied in [4-10] together with the computational power and flexibility of the function smoothing by the Fourier transform (FT) [11]. The description of the clustering data set by a density function is a powerful tool, which provides us with a mathematically more accessible formulation of the data set. \\
Recently, we developed a Fourier-based algorithm, which works very well as a smoothing approach for supervised learning [12]. In this work, the idea of applying the Fourier transform for model training and smoothing the corresponding function will be applied to the density function represented by the Dirac mixture distribution to train a model describing the cluster data. However, there are essential differences between the supervised and the unsupervised approaches because, in our unsupervised algorithm, we do not have any known response, which would use as training data. Thus, we use a different approach based on the convolution convergence between the original and the smoothed density function to determine the optimal model. In the next section, we explain in-depth the mathematical formulation of the problem and how the algorithm works.

\vskip 0.1in

\section{Clustering Optimization by a Fourier-based Algorithm }
\label{sec:headings}
\ Consider a data set $\{(x_i,y_i): i=1,2,...,N\}$ with $N$ distinct data points and two predictor variables $x$, $y$. Mathematically, we can define a corresponding density function by the Dirac mixture Distribution as follows:

\begin{equation}
\ \rho(x,y)=\sum_{i=1}^{N}\delta(x-x_i,y-y_i),
\end{equation}

where $\delta(x-x_i,y-y_i)$ is the Dirac delta function [13] located at the point $(x_i,y_i)$ and $\rho(x,y)$ is the density function, which contents the complete available information (positions) of the sampled data set. The meaning of the density function can be easily understood if we integrate it over an arbitrary space subset $\Omega \in \rm I\!R^2$ ($\rm I\!R$ the set of real numbers) as follows:

\begin{equation}
\ \iint \limits_{\Omega}dxdy\rho(x,y)=\sum_{i=1}^{N}\iint \limits_{\Omega}dxdy\delta(x-x_i,y-y_i)=N_{\Omega},
\end{equation}

where $N_{\Omega}$ is the number of the data points included in the ${\Omega}$. The density function $\rho(x,y)$ results from $N$ sampled data points, thus it is just a rough estimate of the original density function. Basically, the smoothness constraint must be imposed on the $\rho(x,y)$, which is valid in many real-world problems. Consequently, the spectrum of the $\rho(x,y)$ should be filtered by a Gaussian filter [14]. The spectrum of the density function is the result of applying the Fourier transform (FT) on the $\rho(x,y)$:
\begin{equation}
\ \tilde{\rho}(f_x,f_y)=\mathcal{F}\{{\rho}(x,y)\}=\iint \limits_{x,y=-L/2}^{x,y=+L/2}dxdy\rho(x,y)e^{-2\pi i(x f_x+y f_y)},
\end{equation}

where $\mathcal{F}$ is the FT operator and $(f_x,f_y)$ is the frequency pair. The $L$ is the width of the spatial space enclosing the sampled data set ${(x_i,y_i)}$. Multiplying the spectrum of the density function $\tilde{\rho}(f_x,f_y)$ by a Gaussian distribution with the standard deviation $\tilde{\sigma}$ results in the spectrum of the smoothed density function $\tilde{\rho}_s(f_x,f_y)$:
\begin{equation}
\ \tilde{\rho}_s(f_x,f_y)=\tilde{\rho}(f_x,f_y)  \frac{1}{2\pi \tilde{\sigma}^2}e^{- (f_x^2+f_y^2)/2 \tilde{\sigma}^2}.
\end{equation}
The smoothed density function ${\rho}_s$ is calculated by applying inverse FT (IFT) on Eq. (4):
\begin{equation}
\ {\rho}_s(x,y)=\mathcal{F}^{-1}\{\tilde{\rho}_(f_x,f_y)  \frac{1}{2\pi \tilde{\sigma}^2}e^{- (f_x^2+f_y^2)/2 \tilde{\sigma}^2}\},
\end{equation}
where $\mathcal{F}^{-1}$ is the IFT operator. Equation (5) can be converted to an equivalent equation in the spatial space by the use of the convolution theorem [15]:
\begin{equation}
\ {\rho}_s(x,y)={\rho}(x,y)*e^{-2\pi^2 \tilde{\sigma}^2(x^2+y^2)},
\end{equation}

where $*$ is the convolution operator. As can be seen in Eq. (6), the IFT of the Gaussian distribution is again a Gaussian function with a standard deviation $\sigma=1/(2\pi \tilde{\sigma})$. In principle, the convolution of the density function with the Gaussian function is responsible for the desired smoothness.
From plugging Eq. (1) in Eq. (6) and considering the definition of the Dirac Delta function follows:
\\
\begin{equation}
\ {\rho}_s(x,y)=\sum_{i=1}^{N}\delta(x-x_i,y-y_i)*e^{-2\pi^2 \tilde{\sigma}^2(x^2+y^2)}=\sum_{i=1}^{N}e^{-2\pi^2 \tilde{\sigma}^2[(x-x_i)^2+(y-y_i)^2]},
\end{equation}

which shows, that basically, the smoothed density function is a summation of $N$ Gaussian functions (with the standard deviation $\sigma$) placed at the $N$ data points $(x_i,y_i)$. However, Eq. (7) computationally is very inefficient, thus, in the algorithm, the $\rho_s(x,y)$ will be obtained by the use of FT and IFT in Eqs. (3) and (5), which are much more efficient: 

\begin{equation}
\ {\rho}_s(x,y)=\frac{1}{2\pi \tilde{\sigma}^2}\mathcal{F}^{-1}\{\mathcal{F}\{\sum_{i=1}^{N}\delta(x-x_i,y-y_i)\}  e^{- (f_x^2+f_y^2)/2 \tilde{\sigma}^2}\},
\end{equation}

The Equation (8) is our estimate to the original density function corresponding to the total data. The standard deviation $\tilde{\sigma}$ is the most significant unknown parameter, which still must be found by the algorithm.
Due to the applying the discrete Fourier transform (DFT) [16] in the algorithm, the minimum spacing in the frequency space $df$ is $1/L$, thus, it is also used as the minimum value of the standard deviation in the frequency space. Consequently, the start value for the $\tilde{\sigma}$ is $1/L$  and its value in the $n$-th iteration is defined as follows:
\begin{equation}
\ \tilde{\sigma}_n=\frac{n}{L}.
\end{equation}

The algorithm should find the best $\tilde{\sigma}_n$ value for the optimal model. Clearly, for very low values of the $\tilde{\sigma}_n$, the number of significant frequencies contributing to the density function would be too low, thus the density function could not represent the significant variations in the sampled data, which are responsible for resolving the clusters. In such cases, some or all clusters would merge together and create bigger clusters or one single cluster. In other words, the corresponding model would be too simple and underfit. In contrast, overfitting the model could happen if it includes too high frequencies, which are basically responsible for extreme fine variations in the data distribution. However, resolving the too fine structures is not desirable because the clusters would split into insignificant tiny sub-clusters, which actually should belong to one single unified cluster. To determine the best $\tilde{\sigma}_n$, the corresponding model will be evaluated by the correlation value between the ${\rho}(x,y)$ and ${\rho}_s(x,y)$. However, as may be expected, it increases monotonically with $\tilde{\sigma}_n$. But if we consider its changes, the influence of the underfitting and overfitting may compensate for a $\tilde{\sigma}_n$. Thus, the algorithm stops where the difference in the correlation in two successive iterations reaches the minimum value of $\epsilon$:
\begin{equation}
\mid corr(\rho_s,\rho_s^{(n)})-corr(\rho_s,\rho_s^{(n-1)})\mid< \epsilon.
\end{equation}
Inequality (10) is the convergence criterion for the algorithm and the $\epsilon$ is the convergence parameter, which practically is highly independent of the data sets and it can be fixed and for the most applications set to the value 0.01.\\
The FT and IFT in Eq. (8) will be accomplished by the use of a fast Fourier transform (FFT) [17], which is a fast algorithm for implementation of the DFT. The standard DFT requires a uniform sampled data, however, in general, the given sampled data points $\{(x_i,y_i)\}$ are randomly distributed. Therefore, they must be mapped to an equidistant mesh. Consider the sets of both predictor values $\{x_i\}$ and $\{y_i\}$, which are sorted in ascending order. The distance between $i$-th and $(i+1)$-th successive points in each set can be easily calculated:
\begin{equation}
dx_i=x_{i+1}-x_i \quad , \quad dy_i=y_{i+1}-y_i.
\end{equation}
Now we take the $M$ ($<1M<N$) first data points and define the minimum of the mean values corresponding to each predictor as the spatial spacing of both predictors for the equidistant mesh:
\begin{equation}
dx=dy=min(\frac{1}{M} \sum_{i=1}^{M}dx_i,\frac{1}{M} \sum_{i=1}^{M}dy_i).
\end{equation}

The reason for the condition $1<M<N$ is the fact that the spacing must theoretically be sufficiently small to prevent deviations between the original predictor values and their new values caused by the mapping to the equidistant mesh. However, too small values of the spacing increase the number of the mesh points and consequently increase the computational complexity of the FFT.
In many cases, an appropriate value for $M$ is about $5\%$ of the $N$. Now, each data point $\{(x_i,y_i)\}$ will be mapped to the nearest mesh point. In principle, it is probable, that some data points have to be mapped to the same mesh point. The mesh points actually are the centers of the pixels with the area $dxdy$ and each pixel represents one single data point like a data point in the continuous space. Thus, for each mesh point, only one of the data points will be considered. It must be mentioned, that the Dirac delta function in the discrete form will be converted to the Kronecker delta function [16]. Consequently, the density value for pixels including a data point will be one and for other points is zero.\\
In principle, the locations of the local maximum points of the calculated smoothed density function ${\rho}_s(x,y)$ are the estimates for the desired cluster positions. To find the local maxima points, the space with the width and length $L$ will be segmented into tiny squares. If the segments are enough small that there is solely one maximum point inside them, the maximum point can be easily found by comparing the values of the ${\rho}_s(x,y)$ in each segment. If the maxima points do not lie on the boundary they are local maxima and their locations must be saved. To minimize the computational complexity, it is required to minimize the number of segments, which means maximizing the size of the segments. The maximum size of the segments can be estimated by the fact that there is a lower limit for the distance between the local maxima. As shown in Eq. (7), the ${\rho}_s(x,y)$ is composed of many Gaussian functions. Basically two Gaussian functions will merge together if the distance between their maxima is smaller than a critical distance and their superposition will have a single maximum. It can be easily shown, that the critical distance for two identical Gaussian functions with the same standard deviation $\sigma$ is equal to the $2 \sigma$. We want to use this value as an estimate for the minimum distance between the local maxima of the smoothed density function and consequently the minimum required width $w_c$ of the sliding windows is:
\begin{equation}
\ w_c= 2 \sigma_n= \frac{1}{\pi \tilde{\sigma}_n}.
\end{equation}

In principle, some of the local maxima can be on the edges of the sliding windows. To find such points, the sliding windows with three different widths can be applied so that no local maximum point can lie at the same time on the edges of all three window groups. One choice for three window sizes $w_1$, $w_2$ and $w_3$ is:
\begin{equation}
\ w_1 m=w_2(m+1)=w_3(m+2) , \quad w_1,w_2,w_3 \leq w_c,
\end{equation}

The smoothing procedure can cause some invalid local maxima, which typically exhibit very small density values. To exclude such points, too small values of the ${\rho}_s(x,y)$ will be set to zero before the sliding process starts. A reliable estimate of the allowed minimum density value for the local maxima is $0.1$. This parameter will be sufficiently independent of the data if the ${\rho}_s(x,y)$ is shifted so that its minimum is set to zero and it is normalized. 

\section{Results}
\ The proposed algorithm is applied to the data set presented in Fig. 1. Both predictors $x$ and $y$ are normalized. The data is composed of 6 clusters generated by  2-dimensional asymmetric Gaussian distributions with different standard deviations and different numbers of the cluster data points.

\begin{figure}[H]
 \centering
 \includegraphics[width=7cm]{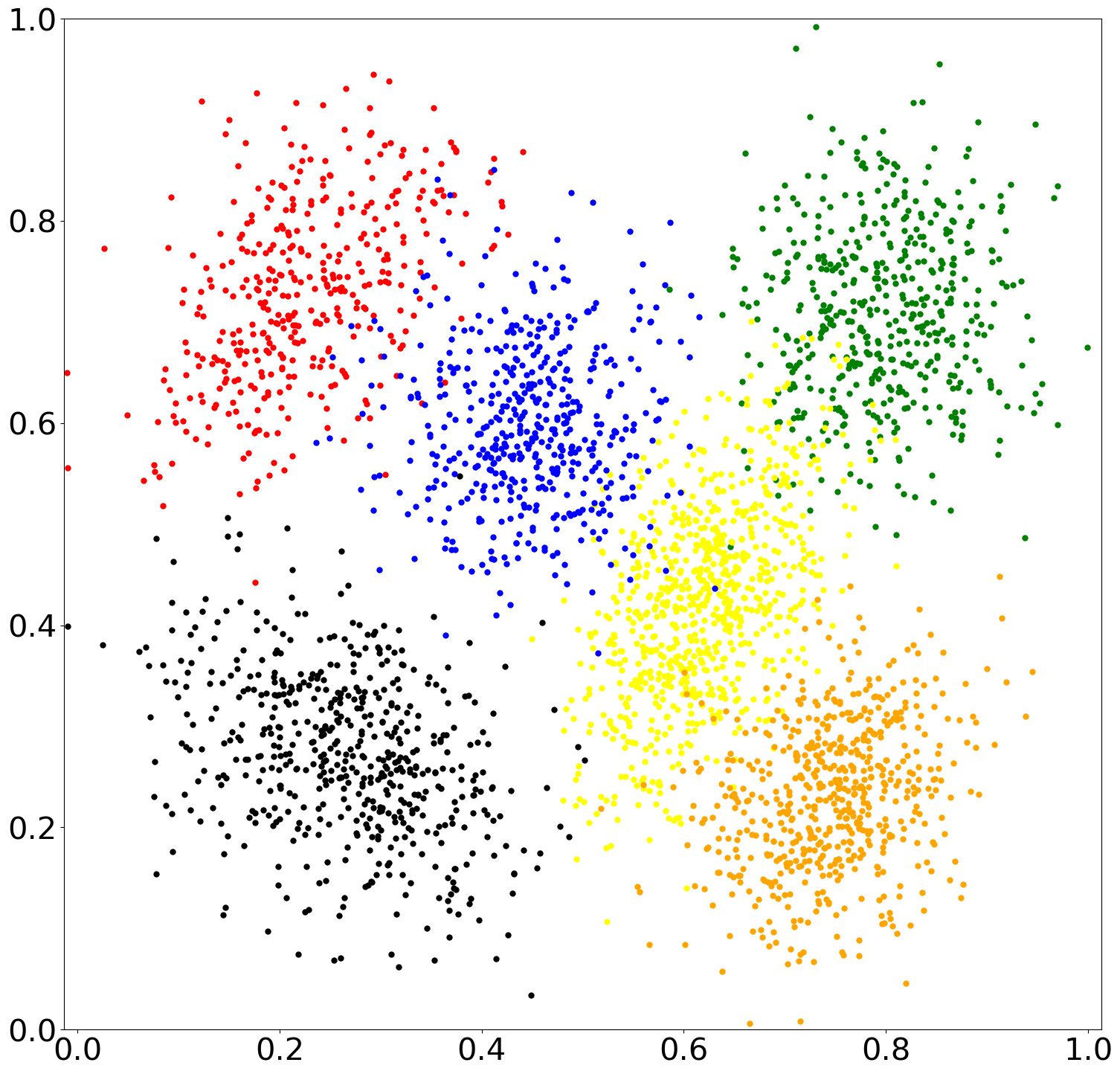}
 \caption{The data set is composed of 6 different clusters related to two normalized predictors $x$ and $y$.}
 \label{fig:soheil3}
\end{figure}

Table 1 shows the coordinates $(\mu_x,\mu_y)$ of the cluster centroids, which have to be found by the use of the algorithm.

\begin{table}[hbt!]
 \caption{Original coordinates of the centroids}
  \centering
  \begin{tabular}{lllllll}
    \toprule
    \  & \ 1 \ & \ 2 \ & \ 3 \ & \ 4 \ & \ 5 \ & \ 6     \\
    \midrule
    $\mu_x$ & 0.26 & 0.22 & 0.80 & 0.62 & 0.44 & 0.75 \\
    $\mu_y$ & 0.27 & 0.73 & 0.71 & 0.42 & 0.60 & 0.23  \\
    \bottomrule
  \end{tabular}
  \label{tab:table}
\end{table}

Figure 2 shows the data set mapped to the equidistant mesh with the computed spacing for the spatial space $dx=dy=0.0033$ and pixel number $N_x=N_y=303$. The total number of the data points related to all clusters is $N_{clsuter}=3350$ and after mapping $154$ data points are removed.  

\begin{figure}[H]
 \centering
 \includegraphics[width=8cm]{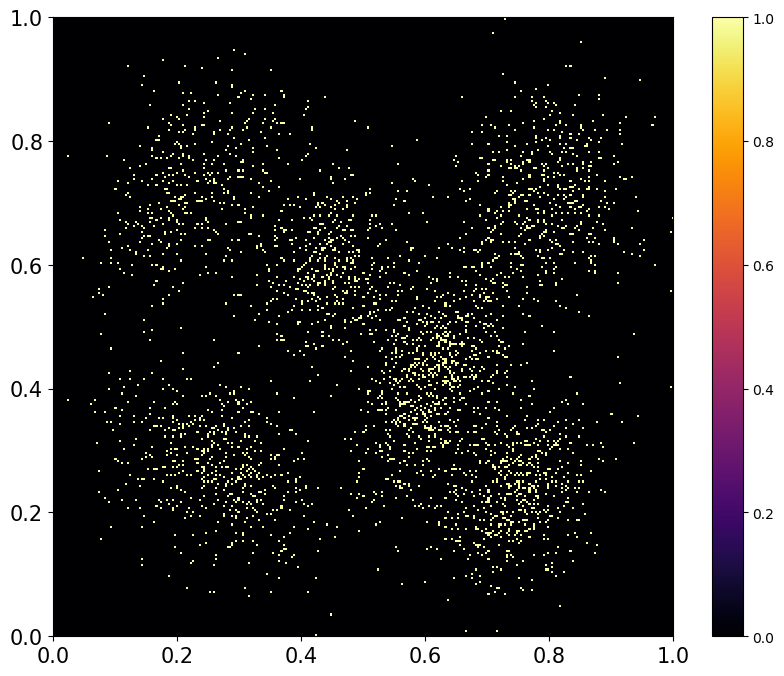}
 \caption{The data set mapped to the equidistant mesh}
 \label{fig:soheil1}
\end{figure}

The computed smoothed density function after removing the invalid local maxima with the parameter 0.1 is shown in Fig. (3). 
The algorithm satisfied the convergence criterion (Inequality (10)) after only $4$ iterations with the convergence parameter $\epsilon=0.01$.

\begin{figure}[H]
 \centering
 \includegraphics[width=8cm]{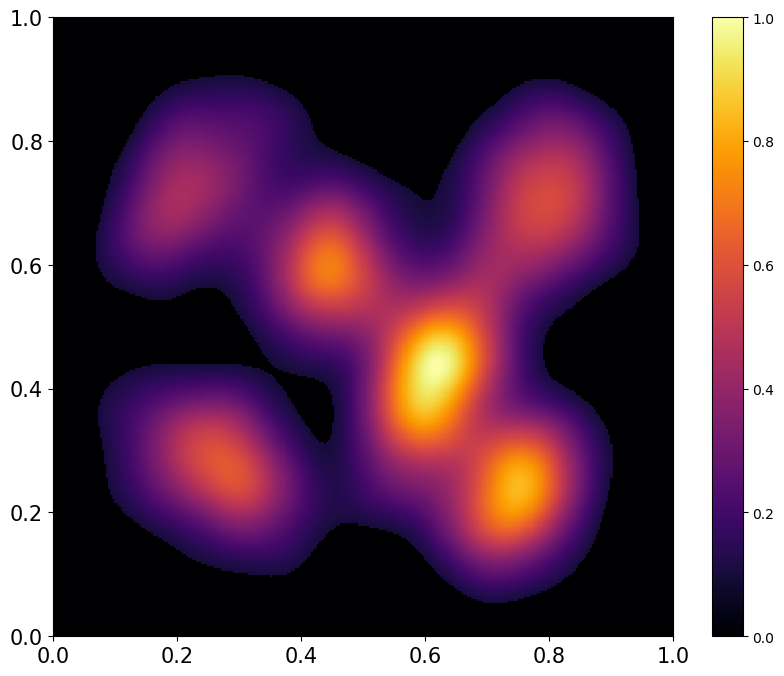}
 \caption{Smoothed density function}
 \label{fig:soheil2}
\end{figure}

The sliding windows with three different sizes $w_1=0.077, w_2=0.071, w_3=0.067 $ have been applied to localize the local maxima. As presented in Table 2, exactly $6$ local maxima are found and their computed locations remarkably conform to the original values shown in Table 1. To evaluate the algorithm performance and its accuracy in the finding the cluster centroids, the root-mean-square error (RMSE) [18] between the values in Table 1 and 2 is calculated. The RMSE value is only $0.012$, which demonstrates the reliability of the proposed algorithm.
\\
\\
\\
\\

\begin{table}[hbt!]
 \caption{Computed coordinates of the centroids}
  \centering
  \begin{tabular}{lllllll}
    \toprule
    \  & \ 1 \ & \ 2 \ & \ 3 \ & \ 4 \ & \ 5 \ & \ 6     \\
    \midrule
    $\mu_x$ & 0.28 & 0.21 & 0.79 & 0.62 & 0.44 & 0.75 \\
    $\mu_y$ & 0.27 & 0.71 & 0.70 & 0.44 & 0.59 & 0.24  \\
    \bottomrule
  \end{tabular}
  \label{tab:table}
\end{table}

\section{Conclusion}

We presented a Fourier-based algorithm for finding the number and locations of clusters. The algorithm can be used as a pre-processing step in the K-Means and Gaussian mixture method and in general for each clustering algorithm, which requires a priori information concerning the number of clusters or initialization of cluster centroids. The algorithm converges very fast and it accurately computes the position and number of the clusters. By the use of the developed method, the typical initialization problem in the K-Means method can be solved. Because of the use of a fast Fourier algorithm, the computational complexity of the algorithm is very low. The only used parameters in the algorithm are convergence criterion, the parameter for excluding the invalid local maxima and the parameter, which is used to determine the minimum spacing in the spatial space. However, practically both first parameters are insensitive to the data and with very high reliability both parameters can be fixed to the proposed values. The parameter related to the mesh generation is fundamentally not inevitable but it can easily set to the minimum calculated spacing of both predictors. In other words, it is only required to reduce computational time, however, the proposed value of $5\%$ can be used without any significant dependence on the data sets.  Without loss of generality, three parameters can be considered as rather invariant parameters. This unique property reasonably protects the algorithm against fatal errors related to the parameter choice. To our view, the accuracy and overall feasibility of the algorithm demonstrate that it can be applied in clustering problems for finding the number of clusters and as a powerful general solution for initialization issue in clustering algorithms.

\bibliographystyle{unsrt}  


\end{document}